# A Spatial Guided Self-supervised Clustering Network for Medical Image Segmentation

Euijoon Ahn[1], Dagan Feng[1,2] and Jinman Kim[1]

[1] School of Computer Science, The University of Sydney, Australia
[2] Med-X Research Institute, Shanghai Jiao Tong University, China
euijoon.ahn@sydney.edu.au

**Abstract.** The segmentation of medical images is a fundamental step in automated clinical decision support systems. Existing medical image segmentation methods based on supervised deep learning, however, remain problematic because of their reliance on large amounts of labelled training data. Although medical imaging data repositories continue to expand, there has not been a commensurate increase in the amount of annotated data. Hence, we propose a new spatial guided self-supervised clustering network (SGSCN) for medical image segmentation, where we introduce multiple loss functions designed to aid in grouping image pixels that are spatially connected and have similar feature representations. It iteratively learns feature representations and clustering assignment of each pixel in an end-to-end fashion from a single image. We also propose a context-based consistency loss that better delineates the shape and boundaries of image regions. It enforces all the pixels belonging to a cluster to be spatially close to the cluster centre. We evaluated our method on 2 public medical image datasets and compared it to existing conventional and self-supervised clustering methods. Experimental results show that our method was most accurate for medical image segmentation.

## 1 Introduction

Supervised deep learning methods allow the derivation of image features for a variety of image analysis problems using underlying algorithms and large-scale labelled data [1]. In the medical imaging domain, however, there is a paucity of labelled data due to the cost and time entailed in manual delineation by imaging experts, inter- and intra-observer variability amongst these experts and then the complexity of the images themselves where there may be many different appearances based on, for instance, bone and soft tissues windows on computed tomography (CT) and different sequences on Magnetic Resonance (MR) and noise on Ultrasound (US) images.

Researchers have employed many different approaches to help solve these challenges including deep learning with transferable knowledge across different domains and fine-tuning those knowledges with a relatively smaller amount of labelled image data (i.e., domain adaptation). Other approaches use unsupervised feature learning [2, 3] where the aim is to learn invariant local image features using algorithms such as sparse coding and auto-encoder. Recently, self-supervised learning, a form of unsupervised learning, where the data themselves generate supervisory signals for the feature learning, has shown great success in many computer vision [4] and medical image segmentation tasks [5-7]. For example, Zhuang et al. [8] trained a convolutional neural network (CNN) in a self-supervised manner by predicting the spatial transformation of 3D CT scan images for brain tumour segmentation. Similarly, Tajbakhsh et al. [9] constructed supervisory signals by predicting colour, rotation and noise for lung lobe segmentation in CT scans. Another approach to construct a supervisory signal is to use image clustering [10-12]. The key concept of self-supervised clustering is to use clustering assignments (i.e., cluster labels) as surrogate labels to learn the parameters of CNNs. Caron et al. [10] proposed a DeepCluster that iteratively learns and improvs the feature representation and clustering assignment of image features during CNN training. Similarly, Ji et al. [11] improved the clustering by maximising the mutual information between spatially transformed image patches (Invariant Information Clustering (IIC)). In medical imaging, Moriya et al. [13] used $k$-means clustering to group similar pixels on micro-CT images and learned the feature representation of pixels using the cluster labels. While these self-supervised clustering methods have shown to be effective in various medical image segmentation tasks, they are limited by the manual selection of cluster size (e.g., $k$ from $k$-means) and may fail when there are complicated regions with fuzzy boundaries, various shapes, artifacts and noise.

In this paper, we propose a new spatial guided self-supervised clustering network (SGSCN) for medical image segmentation, where we iteratively learn optimal cluster size and improve the feature representation of each pixel. We also design a context-based consistency loss as a differentiable loss function that aids in segmenting image regions with fuzzy boundaries and noise. The context-based consistency loss enforces all the pixels belonging to a cluster to be spatially close to the cluster centre. We validated our approach on 2 public datasets and compared it to other unsupervised clustering and self-supervised clustering methods.

## 2 Materials and Methods

### 2.1 Materials

We used 2 public datasets for our experimental analysis. Each dataset was used to access the performance of our method on 2 different problems.

**Skin lesion segmentation** – we used PH2 [14] public dataset. It provides 200 dermoscopic image studies, including 80 common nevi, 80 atypical nevi and 40 melanomas. Manually annotated lesions from expert dermatologists were available from the PH2 and used as the ground truth data. The PH2 dataset provides various images with complex skin conditions.

**Liver tumour segmentation** – we used Sun Yat-sen University US (SYSU-US) public dataset [15]. It provides 20 sets of 2D US image sequences containing 10-30 images of abdomen with liver tumour. The ground truth images were annotated by experts and provided by the SYSU. A subset of 100 US images were used for the evaluation of our method on single slice image segmentation and this was done by randomly selecting 5 US images from each image sequence, consistent with other research [15].

### 2.2 Overview of the SGSCN

A schematic of our method is shown in Fig. 1. Given an image $x_n$, we train a convolutional segmentation network $F$ parameterized by $\theta_f$ that generates segmentation map $S_n = F(x_n; \theta_f)$. We then normalise the segmentation map $S_n$ such that $\hat{S}_n$ has zero mean and unit variance. The final segmentation map (cluster label $C_n$ of each pixel) is obtained by selecting the channel-wise dimension that has maximum response value in $\hat{S}_n$ (i.e., *argmax* function). The parameters $\theta_f$ were trained by minimizing the cross-entropy loss between $\hat{S}_n$ and $C_n$. Sparse spatial loss and context-based consistency loss are also used to enhance the clustering assignments by understanding the spatial relationships of image pixels and regions. The SGSCN iteratively learns and improves feature representation and clustering assignment of each pixel in an end-to-end fashion.

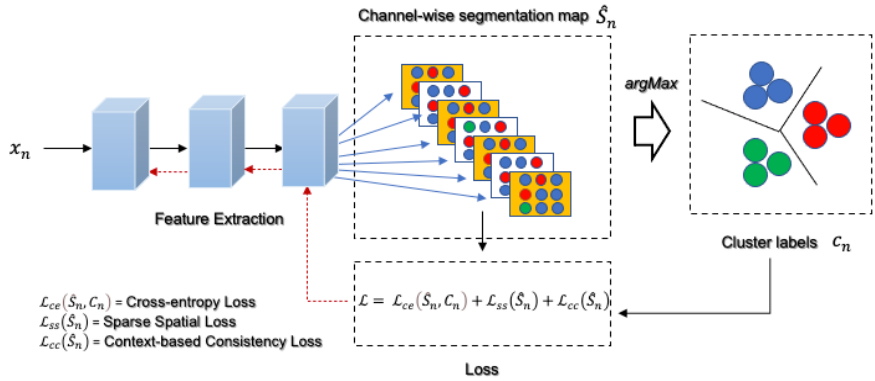

**Fig. 1.** A schematic of our SGSCN.

### 2.3 Cross-entropy Loss for Self-supervised Clustering

As in a standard CNN, the parameters of the convolutional segmentation network $F$ are learned by calculating the cross-entropy loss between $\hat{S}_n$ and $C_n$, which is defined as follows:

$$\mathcal{L}_{ce}(\hat{S}_n, C_n) = \sum_{n=1}^{N} \sum_{p=1}^{i} -\delta(p - C_n) \ln \hat{S}_{n-1} \tag{1}$$

where $p$ is the cluster index ($p = 1, \cdots, i$) and $\delta(\cdot)$ is the indicator function that assigns the value 1 when $(p - C_n)$ is equal to 0 and the value 0 for otherwise. The cross-entropy loss is minimized using stochastic gradient descent (SGD).

### 2.4 Sparse Spatial Loss

Classic cross-entropy loss ignores the spatial relationships of image pixels or regions, which generate sub-optimal parameters during CNN training. Unfortunately, this makes it difficult to group spatially connected pixels (e.g., edges) or regions.

$L_1$-norm based regularization techniques are used widely in image restoration and denoising and has proven effectiveness when dealing with sparse data [16, 17]. It has also shown to less penalize subtle edges or spatially connected regions in images. Thus, we use the $L_1$-norm to measure the vertical and horizontal differences of the segmentation map $\hat{S}_n$ to better understand the spatial relationships of image pixels and regions. This is defined as follows:

$$\mathcal{L}_{ss}(\hat{S}_n) = \sum_{k=1}^{W-1} \sum_{l=1}^{H-1} \left\| \hat{S}_{k+1,l} - \hat{S}_{k,l} \right\|_1 + \left\| \hat{S}_{k,l+1} - \hat{S}_{k,l} \right\|_1 \qquad (2)$$

where $W$ and $H$ are the width and the height of the input image and $\hat{S}_{k,l}$ denotes the pixel value at $(k, l)$ coordinate in the segmentation map $\hat{S}_n$.

### 2.5 Context-based Consistency Loss

Our context-based consistency loss enforces pixels belonging to a cluster to be spatially close to the cluster centre. The cluster centre of $C_n$ along with axis $k$ can be calculated by transforming each cluster as a spatial probability distribution function as follows:

$$C_n^k = \sum_n \sum_{k,l} k \cdot \hat{S}_n(k,l) / \sum_{k,l} \hat{S}_n(k,l). \qquad (3)$$

Using the cluster centres, we then calculate the context-based consistency loss according to:

$$\mathcal{L}_{cc}(\hat{S}_n) = \sum_n \sum_{k,l} \|(k,l) - (C_n^k - C_n^l)\|^2 \cdot \hat{S}_n(k,l) / \sum_{k,l} \hat{S}_n(k,l). \qquad (4)$$

This loss function is used for penalizing pixels that are spatially located away from the cluster centre.

### 2.6 Training via Backpropagation

The overall loss function for the convolutional segmentation network $F$ is the sum of the cross-entropy loss, sparse spatial loss and context-based consistency loss. We set an arbitrary maximum number of possible clusters $i$ (as in Eq. 1) and this is iteratively minimised using the overall loss function. Similar image pixels would be assigned to same clusters during the course of training, making the unique number of cluster smaller than initially defined maximum cluster size. This process is repeated until the clustering and the loss become stable.

## 3   Experimental Setup

### 3.1   Evaluation

We evaluated our method by comparing it to other unsupervised and self-supervised clustering methods. As the baseline, we compared our method to well-established unsupervised *k*-means clustering algorithm. We also compared it to the state-of-the-art self-supervised clustering methods – DeepCluster [10] and IIC [11]. We used the three standard metrics including Dice similarity coefficient (DSC), Hammoude distance (HM) and XOR which are routinely used among researchers to assess segmentation performance. A higher DSC score corresponds to a better result. In contrast, lower scores for HM and XOR correspond to better results. Since the background regions, such as dark corners, in dermoscopic images and in US images are also considered as one of largest segments within an image, we only considered the predicted segment (i.e., cluster) that had the largest overlap with the GT segment in our evaluation.

### 3.2   Implementation Details

For our convolutional segmentation network $F$, we adopted a shallow CNN architecture, comprises 3 convolutional layers, each of which has ReLU activation and batch normalisation function. Each convolutional layer has the following network parameters: kernel size of 3 × 3, stride of 1, pad size of 1 and, filter size of 100. Here the filter size of 100 is equivalent to the maximum number of possible clusters $i$ (as defined in Eq. 1). We set the uniform learning rate of 0.1 with a momentum of 0.9 for skin lesion segmentation task and a smaller learning rate of 0.05 with a momentum of 0.9 for liver tumour segmentation. We trained our network on a GeForce GTX 1080 Ti GPU (11GB memory). We used an empirical process to discover appropriate number of clusters (3 to 8) for *k*-means, DeepCluster and IIC. For *k*-means, $k=3$ had the highest accuracy for skin lesion segmentation and $k=6$ for liver tumour segmentation. We set $k=3$ for DeepCluster and IIC in our all experiments.

## 4   Results

The results of skin lesion segmentation from dermoscopic images are shown in Table 1. Our SGSCN generated higher scores and had the best overall DSC average (83.4%) and the best XOR (28.2%) accuracy. The results of liver tumour segmentation from US images are outlined in Table 2 and they show that our method had the highest DSC value (63.2%), and the best HM (46.2%) and XOR (52.3%) scores. We show the results for four selected images in Fig. 2 with two dermoscopic images in the 1st and 2nd rows and two US images in the 3rd and 4th rows. Fig. 3 shows the relative improvement in segmentation accuracy due to the use of sparse spatial and context-based consistency losses, in addition to the cross-entropy loss. We also show the sample segmentation results of using sparse spatial and context-based consistency loss in Fig. 4.

Table 1. The segmentation results on PH2 (skin lesion segmentation) compared to other methods

| Mean% | *k*-means (*k=3*) | *k*-means (*k=4*) | DeepCluster | IIC | Our Method |
|---|---|---|---|---|---|
| DSC | 71.3 | 67.7 | 79.6 | 81.2 | **83.4** |
| HM | 130.8 | 165.2 | 35.8 | 35.3 | **32.3** |
| XOR | 41.3 | 44.9 | 31.3 | 29.8 | **28.2** |

Table 2. The segmentation results on SYSU-US (liver tumour segmentation) compared to other methods

| Mean% | $k$-means ($k=5$) | $k$-means ($k=6$) | DeepCluster | IIC | Our Method |
|---|---|---|---|---|---|
| DSC | 37.1 | 39.3 | 60.9 | 58.3 | **63.2** |
| HM | 93.2 | 95.2 | 47.9 | 56.2 | **46.2** |
| XOR | 75.1 | 73.3 | 55.2 | 55.8 | **52.3** |

## 5    Discussion

Our findings indicate that our SGSCN a) outperformed other unsupervised and self-supervised clustering methods; b) was able to locate the lesion and its proximity to image boundaries (see Fig. 2); c) progressively improved the feature representation and the clustering assignments of each pixel and d) enhanced the clustering assignment of each pixel using our sparse spatial and context-based consistency loss (see Fig. 3 and Fig. 4).

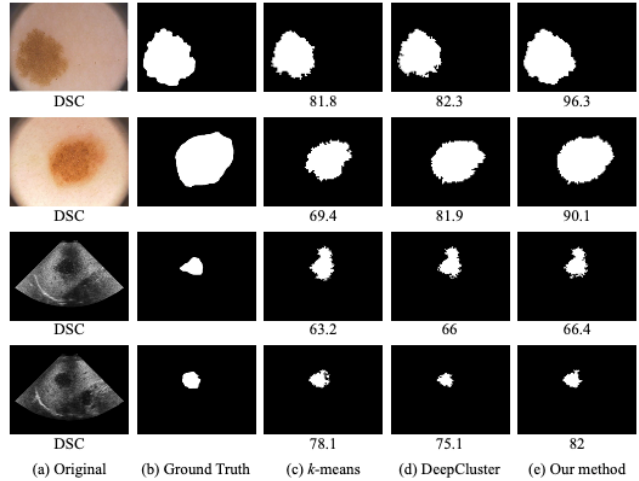

**Fig. 2.** Segmentation results from 4 study examples (top two rows: dermoscopic images and bottom two rows: US images), where (a)-(e) represent the original image in column 1, ground truth in column 2, and the segmentation results from column 3 to column 7 for $k$-means, DeepCluster and our method.

Our SGSCN had higher accuracy than other recent self-supervised clustering methods. This is attributed to our sparse spatial loss that helps in segmenting spatially connected regions in images (see Fig. 4, row1). The context-based consistency loss further enhances the segmentation by focusing on segments that are spatially close to the cluster centre (see Fig. 4, row2).

The quality of clustering used in the conventional unsupervised *k*-means was not as robust as those from recent self-supervised clustering methods such as DeepCluster and IIC. DeepCluster and IIC also iteratively learned feature representation and clustering assignments and hence had a higher accuracy in skin lesion and liver tumour segmentation when compared to *k*-means. Their performance, however, varied widely depending on the initial selection of fixed cluster size and do not incorporate the spatial relationships of image pixels and regions during CNN training.

In the segmentation of skin lesion, our SGSCN outperformed all other unsupervised and self-supervised clustering methods. The DeepCluster and IIC were the next closest to ours in overall DSC and XOR scores. DeepCluster, however, did not consider the spatial relationships of image regions and therefore was not able to accurately segment lesions with fuzzy boundaries (see Fig. 2, row1).

In liver tumour segmentation from US images, our SGSCN outperformed all other self-supervised clustering methods, consistent with the results of skin lesion segmentation. Due to the presence of speckle noise and low contrast in US images, it was more challenging to localise and segment tumour regions. As a consequence, the overall performance of all the methods was reduced when compared to the performance of skin lesion segmentation. Nevertheless, our method still achieved the best DSC and the lowest HM and XOR (see Table 2).

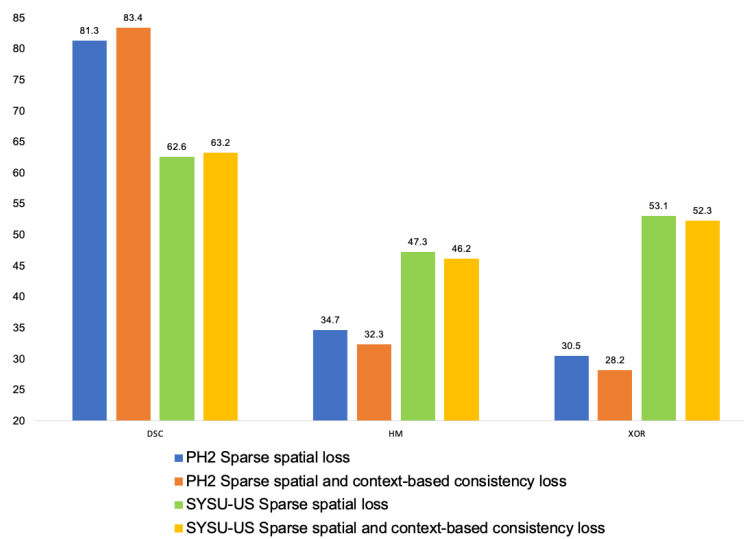

**Fig. 3.** DSC, HM and XOR scores of our SGSCN with sparse spatial loss and our SGSCN with both sparse spatial and context-based consistency loss.

Although our method improved the segmentation of skin lesions and liver tumours in a self-supervised manner, it has some limitations when the lesion or tumour region is very small, not visually distinctive and has incomplete boundary. It is possible that other spatial or geometric constraints such as affine or thin plate spline grid may further improve the medical image feature representation, and we will investigate this as part of our future work.

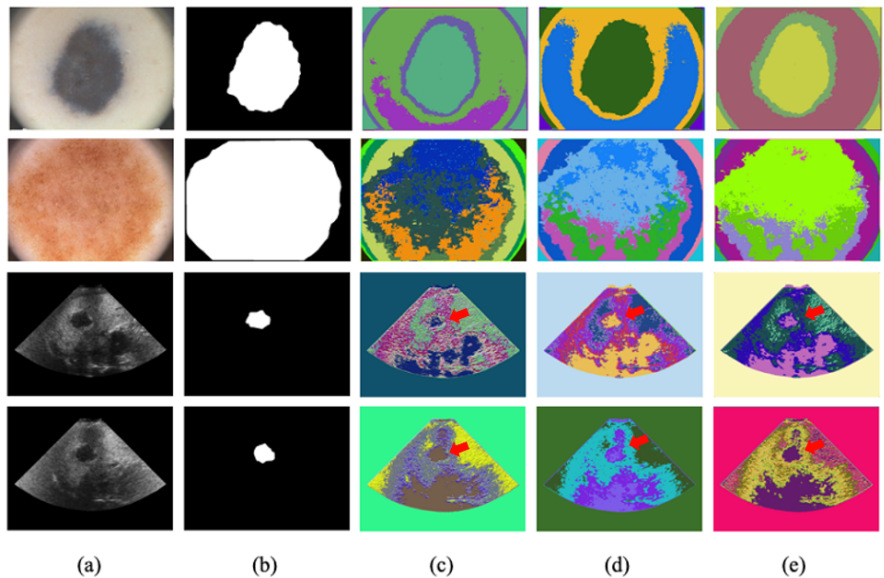

**Fig. 4.** Sample segmentation results of dermoscopic and US images using SGSCN, where (a)-(e) represent the original image in column 1, ground truth in column 2, and the clustering results from column 3 to column 5 for our segmentation network with cross-entropy loss, our segmentation network with cross-entropy and sparse spatial loss and our segmentation network with cross-entropy, sparse spatial and context-based consistency loss.

## 6 Conclusion

In this work, we developed a self-supervised clustering network that iteratively learns image feature representation and clustering of image pixels by characterising the spatial relationships of image regions. We compared our method to other unsupervised and self-supervised clustering methods on 2 public datasets and showed that our method outperformed other methods. Our findings indicate that our method can be applied to various medical image data.

international conference of the IEEE engineering in medicine and biology society (EMBC), pp. 5437-5440. IEEE, (2013)
15. Lin, L., Yang, W., Li, C., Tang, J., Cao, X.: Inference with collaborative model for interactive tumor segmentation in medical image sequences. IEEE transactions on cybernetics 46, 2796-2809 (2015)
16. Fu, H., Ng, M.K., Nikolova, M., Barlow, J.L.: Efficient minimization methods of mixed l2-l1 and l1-l1 norms for image restoration. SIAM Journal on Scientific computing 27, 1881-1902 (2006)
17. Zhang, M., Desrosiers, C.: High-quality image restoration using low-rank patch regularization and global structure sparsity. IEEE Transactions on image Processing 28, 868-879 (2018)